\documentclass[conference,11pt]{IEEEtran}
\IEEEoverridecommandlockouts

\usepackage[T1]{fontenc}
\usepackage{graphicx}
\usepackage{amsmath}
\usepackage[listofformat=parens, justification=centering, subrefformat=subparens]{subfig}
\usepackage{tabularx}
\usepackage{hyperref}
\usepackage{fancyhdr}

\hypersetup{
colorlinks,%
citecolor=black,%
filecolor=black,%
linkcolor=black,%
urlcolor=black
}

\usepackage{comment}
\newcommand\etal{\textit{et\,al.}}
\newcommand\ie{\textit{i.\,e.}}
\newcommand\eg{\textit{e.\,g.}}

\begin{document}

\title{Refining Tuberculosis Detection in CXR Imaging: Addressing Bias in Deep Neural Networks via Interpretability}


\author{\IEEEauthorblockN{1\textsuperscript{st} Özgür Acar Güler}
\IEEEauthorblockA{\textit{Department of Informatics} \\
\textit{University of Zurich}\\
Zurich, Switzerland\\
oezgueracar.gueler@gmail.ch}
\and
\IEEEauthorblockN{2\textsuperscript{nd} Manuel Günther}
\IEEEauthorblockA{\textit{Department of Informatics} \\
\textit{University of Zurich}\\
Zurich, Switzerland\\
manuel.guenther@uzh.ch}
\and
\IEEEauthorblockN{3\textsuperscript{rd} André Anjos}
\IEEEauthorblockA{\textit{Idiap Research Institute} \\
Martigny, Switzerland \\
andre.anjos@idiap.ch}
}

\makeatletter

\def\ps@IEEEtitlepagestyle{%
  \def\@oddfoot{\mycopyrightnotice}%
  \def\@evenfoot{}%
}
\def\mycopyrightnotice{%
  {}
  \gdef\mycopyrightnotice{}
}

\makeatother

\maketitle              

{
  \chead{\footnotesize This is a pre-print of the paper presented at the European Workshop on Visual Information Processing (EUVIP) 2024.}
  \lhead{}
  \thispagestyle{fancy}
}

\begin{abstract}
Automatic classification of active tuberculosis from chest X-ray images has the
potential to save lives, especially in low- and mid-income countries where
skilled human experts can be scarce. Given the lack of available labeled data
to train such systems and the unbalanced nature of publicly available
datasets, we argue that the reliability of deep learning models is limited,
even if they can be shown to obtain perfect classification accuracy on the test
data. One way of evaluating the reliability of such systems is to ensure that
models use the same regions of input images for predictions as medical experts
would. In this paper, we show that pre-training a deep neural network on a
large-scale proxy task, as well as using mixed objective optimization network
(MOON), a technique to balance different classes during pre-training and
fine-tuning, can improve the alignment of decision foundations between models
and experts, as compared to a model directly trained on the target dataset. At
the same time, these approaches keep perfect classification accuracy
according to the area under the receiver operating characteristic curve (AUROC)
on the test set, and improve generalization on an independent, unseen dataset.
For the purpose of reproducibility, our source code is made available
online.\footnote{\href{https://medai.pages.idiap.ch/software/paper/euvip24-refine-cad-tb/}{\scriptsize https://medai.pages.idiap.ch/software/paper/euvip24-refine-cad-tb/}}
\end{abstract}

\begin{IEEEkeywords}
Computer-aided diagnosis, Tuberculosis, Interpretability, Saliency mapping, Label balancing, Bias
\end{IEEEkeywords}

\section{Introduction}
\label{sec:intro}

Chest radiography (CXR) has been a pivotal tool in diagnosing and managing
tuberculosis (TB) for over a century, while its effectiveness largely depends
on the availability and expertise of human interpreters. Recent advancements in
artificial intelligence, particularly in computer-aided detection (CAD)
software, have revolutionized CXR analysis for TB detection. These software
applications not only automate CXR interpretation for TB but also identify
other non-TB radiographic abnormalities~\cite{qin-2021}. Recognizing the
potential of CAD, the World Health Organization (WHO) in 2021 endorsed a
conditional recommendation for using CAD solutions as a substitute for human
readers in TB screening and triage for individuals aged 15 and
above~\cite{who-2021}. This endorsement underscores the growing importance of
CAD in enhancing the accuracy and efficiency of TB diagnosis, especially in
contexts where skilled human readers are scarce~\cite{geric-2023}.

However, challenges arise in the deployment of CAD software in medical
diagnostics. The core algorithms of these CAD products are often deep neural
networks (DNN), which are perceived as \emph{black boxes} since their
decision-making processes are not transparent or not easily understandable.
This opacity in algorithmic functioning makes it difficult to assess and
validate existing solutions and limits the trust in algorithmic
forecasts~\cite{gilpin-2018}. Saliency mapping techniques can play a key role
in addressing the black-box nature of deep learning-based CAD software for CXR
interpretation. By visually highlighting detected radiological findings, these
techniques can assist radiologists in providing more accurate
diagnoses~\cite{gilpin-2018,lipton-2016}. Commercially available CAD software
applications for TB interpretation typically include saliency map
visualizations, as such approaches have been shown to improve diagnostic
capabilities, particularly in complex cases with multiple
abnormalities~\cite{kundel-1990,rajpurkar-2020}.

The lack of publicly available and well-curated datasets for training DNNs for
CXR interpretation in TB applications hinders the ability to evaluate and
develop models that can be deployed responsibly. Furthermore, the small amount
of specifically annotated data renders it difficult to analyze if existing
models, which claim high accuracy, are actually exploiting spurious biases in the
data instead of more desirable and meaningful clinically explainable factors.

In this context, we pose the question if naively training a classifier on the
largest publicly available dataset for TB classification,
TBX11K~\cite{tbx11k-2020}, containing more than 11'000 images, is sufficient to
develop a model that produces reasonable interpretation traits that resemble
human judgment. We further explore the best approaches to achieve
this using only publicly available data. Our main contribution is methodological: training (or
pre-training) of DNN models while compensating for data imbalances lead to improved
interpretation alignment with humans, while retaining comparable generalization
capabilities on test data.  Concretely, we show that: i) Naively training a DNN
leads to a highly accurate model that latches to undesirable image biases; ii)
Pre-training the weights of a classifier with a related proxy multi-objective
classification task containing far more data reduces the interpretation biases
and, finally, iii) Balancing classes during pre-training and fine-tuning can
further align the final classifier to human expectations.

\section{Related Work}
\label{sec:review}

Before the Coronavirus Disease 2019 (COVID-19) pandemic, TB was responsible for
more deaths per year than any other infectious disease~\cite{who-2023}. Albeit
the alarming scenario, the number of public datasets that can be used for
developing and validating models for CXR interpretation in this context remains
relatively low. Essentially, there exist four public datasets for the
development and evaluation of TB-related CAD tools, three of which are
relatively small, containing only a few hundred images each and are, therefore,
of limited application to the development of DNN models.  The more recent
TBX11K dataset~\cite{tbx11k-2020} contains 11'200 CXR images with corresponding
bounding box annotations for image areas that corroborate the attributed image
classes, which can be used to evaluate both classification and radiological
sign localization.  The authors of the dataset propose the development of two
separate models for each task, with relatively high performance.  Compared to
that work, our proposed workflow only relies on a single model trained
for classification, which can adjacently explain reasoning through saliency maps.

Research in automatic detection of Pulmonary TB from CXR images has progressed,
with DNNs such as convolutional neural networks (CNNs) leading current
advancements, demonstrating high accuracy and strong correlation with
ground-truth labels, typically derived from skin or sputum
tests~\cite{liu-2023}. Despite all improvements, realistic scenarios where CAD
from CXR imaging for TB could be useful are rather different from lab
conditions~\cite{who-cxr-2016}. In high-burden countries, for example, TB must
be screened against the general population, with individuals potentially
presenting various other (pulmonary) diseases. Patients with positive skin or
sputum tests, in different stages of the disease, or due to other
co-morbidities (\eg~HIV-positive), may not present classical TB symptoms
clearly visible on CXR images.  Therefore, from a (human) interpretation point
of view, one may argue that CAD for TB, based exclusively on CXR images, should
be limited to identifying factual and reportable radiological signs that can be
detected on the original CXR images. The availability of reproducible
baselines also remains low, as typical datasets used to produce published
results are not released due to privacy restrictions that are especially
prominent in medical data.

Saliency mapping techniques serve as post-hoc interpretative tools that
elucidate the decision-making process of a model, thereby enhancing user trust
and understanding. While these maps do not reveal the intricate internal
workings of a model, they contribute to its interpretability by highlighting
key areas in the input data that influence the model's outcomes. This feature
is particularly beneficial in applications like medical diagnosis where it
helps professionals by revealing biases and crucial decision-influencing
factors. However, the degree of usefulness varies across different saliency
mapping techniques, necessitating a detailed examination of selected methods to
understand their specific contributions and limitations~\cite{score-cam-2020}.

\newcolumntype{C}{X<\centering}
\begin{table*}
    \normalsize
    \caption{\textsc{Models}. This table lists training details of our models.
    Pre-training was performed on the \emph{proxy} dataset NIH-CXR14, while
    fine-tuning was performed on the \emph{target} dataset TBX11K.}
    \label{tab:models}

    \centering
    \begin{tabularx}{.95\textwidth}{|C||C|C|C|C|C|}\hline
        Model &           $M_U$    &  $M_B$   &  $M_{U,U}$ & $M_{U,B}$  & $M_{B,B}$\\\hline\hline
        Pre-training &     ---     &   ---    & unbalanced & unbalanced & balanced\\\hline
        Fine-tuning &   unbalanced & balanced & unbalanced & balanced   & balanced\\\hline
    \end{tabularx}
\end{table*}

Given the shortage of public data for training DNNs to perform classification
of CXRs for TB, it should be possible to re-use similar proxy tasks with far
more data, to pre-train these models before fine-tuning on existing, smaller TB
datasets. For chest X-ray (CXR) data, the NIH-CXR14
dataset~\cite{rajpurkar-2018} can be useful as such a proxy dataset since it
represents the largest publicly available repository to date, encompassing more
than 112,000 images containing up to 14 different labels extracted from
radiology reports. These labels are highly unbalanced and do not align with
signs that are typically related to pulmonary TB onset, but pre-training on
this task can still help our final task. Fine-tuning this model involves
adjusting the weights of the pre-trained network to make it more suited for our
specific task. This can be done by retraining the entire network or just the
final layers.


Rudd~\etal~\cite{moon-2016} addressed multi-label imbalance in DNNs and
introduced the mixed objective optimization network (MOON). MOON re-weights losses for highly imbalanced datasets, effectively
balancing training across classes for each objective individually.\footnote{Objectives in \cite{moon-2016} were to classify the presence of different attributes in facial images.} Compared to other resampling strategies like over- or undersampling, this method refrains from introducing additional biases by retaining the inherent correlations between labels and avoiding information loss in case of undersampling, while also being more efficient during training than oversampling. Weights
$w_i^+$ and $w_i^-$ are adapted based on the counts of samples in the positive
$S^+_i$ and negative class $S_i^-$ for each binary objective $i$:
\begin{align}\hspace*{-1em}\label{eq:balanced-weights}
w_i^+ &=
\begin{cases}
1 & \text{if } S^{-}_{i} > S^{+}_{i} \\
\frac{S^{-}_{i}}{S^{+}_{i}} & \text{otherwise}
\end{cases} &\hspace*{-1em}
w_i^- &=
\begin{cases}
1 & \text{if } S^{+}_{i} > S^{-}_{i} \\
\frac{S^{+}_{i}}{S^{-}_{i}} & \text{otherwise}
\end{cases}\hspace*{-1em}
\end{align}
In \cite{moon-2016}, these weights were used as probabilities to sample the
objectives for which loss values are back-propagated,
but they can also improve interpretability when
used as loss weights \cite{zhang2024cam} in binary cross-entropy:
\begin{equation}
\label{eq:weighted-BCE}
\mathcal J = -\sum_{i=1}^M w_i^{t_i} \bigl[t_i\log f_i(x) + (1-t_i)\log (1-f_i(x)) \bigr]
\end{equation}
where $i$ iterates all objectives, $t_i$ is the binary ground-truth label, and $f_i(x)$ is the prediction of objective $i$ in sample $x$.

From this setup, we pose the question if it is beneficial to re-use the
partially unrelated NIH-CXR14 dataset to pre-train a DNN system to perform
accurate and interpretable TB readout from CXR.  We hypothesize that: i)
Pre-training with NIH-CXR14 helps improve generalization compared to only
using the largest TB dataset available (\ie~TBX11K); ii) Balancing classes
during training will further remove spurious biases and improve human
interpretability of saliency maps produced by DNNs.

\begin{table*}
    \normalsize
    \caption{\textsc{Results}. AUROC of all models on the test set of the
    \emph{target} TBX11K dataset, and on the entire \emph{external} Shenzhen
    dataset. Additionally, the medians of Proportional Energy on the
    \emph{target} test set are shown, after applying Grad-CAM, HiResCAM, and
    Score-CAM to the last convolutional layer of all 5 DenseNet-121 models.}
    \label{tab:results}

    \centering


    \begin{tabular}{|l||r|r||r|r|r|} \hline
        Metric& \multicolumn{2}{c||}{AUROC} & \multicolumn{3}{c|}{Proportional Energy}\\\hline
              & \footnotesize Target & \footnotesize External & \footnotesize Grad-CAM & \footnotesize HiResCAM & \footnotesize Score-CAM\\ \hline\hline
        $M_U$ & 1.00 & 0.79 & 0.077 & 0.077 & 0.088 \\ \hline
        $M_B$ & 1.00 & 0.73 & \it 0.060 & \it 0.058 & \it 0.007\\ \hline
        $M_{U,U}$ & 1.00 & 0.86 & 0.109 & 0.118 & 0.155\\ \hline
        $M_{U,B}$ & 1.00 & 0.88 & 0.234 & 0.237 & 0.182\\ \hline
        $M_{B,B}$ & 1.00 & 0.88 & \bf 0.295 & \bf 0.301 & \bf 0.326\\ \hline
    \end{tabular}
\end{table*}

\section{Data and Methods}
\label{sec:datamethods}

In this work, we make use of 3 publicly available datasets:
NIH-CXR14~\cite{cxr8-2017}, TBX11K~\cite{tbx11k-2020} and
Shenzhen~\cite{jaeger-2014}.  Our \emph{proxy} dataset NIH-CXR14 contains
112'120 images of 30'805 unique patients provided as 8-bit grayscale images
with a resolution of 1024$\times$1024. The images of the dataset are split into 3 patient-disjoint partitions
for training (98'637), validation (6'350) and testing (4'054). The 14
image-level labels in this dataset are: atelectasis, consolidation,
infiltration, pneumothorax, edema, emphysema, fibrosis, effusion, pneumonia,
pleural thickening, cardiomegaly, nodule, mass and hernia.

The \emph{target} dataset TBX11K is used to either directly train a model from
scratch or fine-tune an NIH-CXR14 pre-trained variant.  This dataset consists
of 11'702 24-bit RGB recordings of CXRs of resolution 512$\times$512. Each
sample is from a unique individual. There are 4 types of labels for existing
samples, stratifying the images into: healthy, sick (non-TB), active TB, and
latent TB.  Here, we only consider healthy (3'800) and active TB cases (630),
splitting existing data into 3 partitions for training (2'767), validation
(706) and testing (957), while preserving healthy/active TB stratification.
Beyond labels, samples in the TBX11K dataset also contain bounding boxes in the
original CXR image where radiological signs corroborate label assignment for TB
cases (and only in those cases).  The precise type of radiological sign is not
further annotated, except for its relationship to active TB or latent TB
infection (\eg~scars from previous active TB sickness).

Our \emph{external} dataset is Shenzhen and consists of 662 8-bit RGB CXR recordings
of variable resolution (up to 3000$\times$3000). The samples are classified as
either healthy (326) or active TB (336). This dataset is not partitioned into
subsets and is only used for evaluating the generalization of our DNN models.

In this work, we use a standard DenseNet-121 model
architecture~\cite{densenet-2017}, for its excellent accuracy in image-related
problems and availability in various deep learning software libraries. This
model architecture has also shown to produce more accurate visualizations in
conjuction with current saliency mapping techniques in TB detection compared to
the architecture proposed by Pasa~\etal~\cite{pasa_efficient_2019}. Before
training, each network is pre-initialized with readily available ImageNet
weights.  Training is done through a stock Adam optimizer with default
parameters, guided by the weighted binary-cross entropy loss
\eqref{eq:weighted-BCE} in both the binary \emph{target} task and the
multi-objective binary \emph{proxy} task, either using \emph{unbalanced
training} with $\forall i:w_i^+ = w_i^- = 1 $ or \emph{balanced training} using
class weights according to \eqref{eq:balanced-weights}. Data augmentations
included horizontal flips with 50\% probability during \emph{proxy}
pre-training and elastic deformation~\cite{simard-2003} with 80\% probability
for the binary \emph{target} task. Fine-tuning followed the same training
technique with the same learning rate ($1 \times 10^{-4}$) as for the NIH-CXR14
pre-training.  Sample importance balancing was performed using the
loss-weighting technique \eqref{eq:weighted-BCE}. In total, five models were
trained, as listed in Tab.~\ref{tab:models}.

To evaluate classification performance of TB vs.~healthy CXR images, we report
the area under the receiver operating characteristic curve (AUROC) on the
TBX11K test set.  To assess generalization, we report AUROC on the
\emph{external} Shenzhen dataset.  The level of interpretability of saliency
maps for a given DNN model is measured through the median Proportional
Energy~\cite{score-cam-2020} of the test set, where we utilize the ground-truth
bounding boxes provided in our \emph{target} dataset.
Saliency maps are produced using Grad-CAM~\cite{selvaraju2017gradcam} as a
baseline for comparison with HiResCAM~\cite{hirescam-2021} and
Score-CAM~\cite{score-cam-2020}, which from the previous experience produce
maps that best correlate with human
interpretability~\cite{explainability-2023}.

\section{Results and Discussion}
\label{sec:results}

\begin{figure*}[t]
    \centering
    \subfloat[$M_U$\label{fig:unw-5-lowest-propeng}]{
        \includegraphics[width=\linewidth]{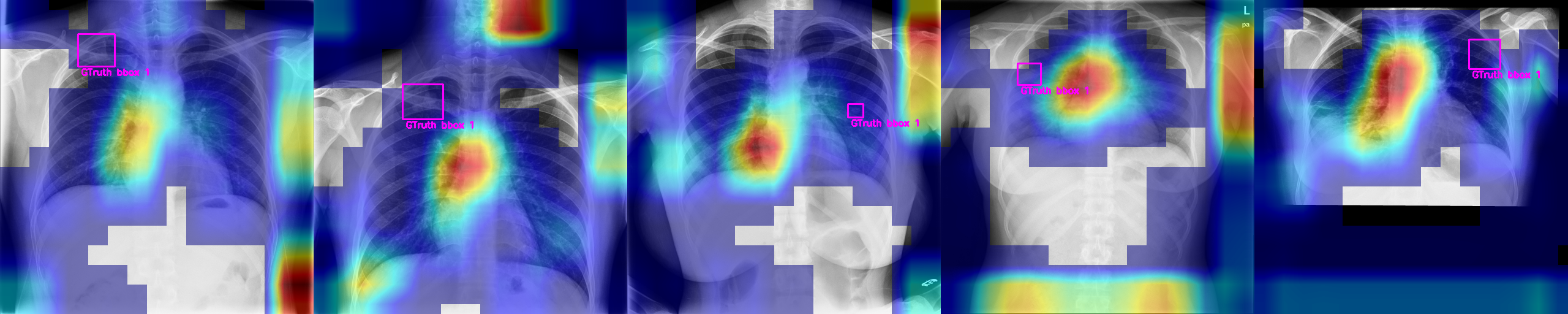}
    }

    \subfloat[$M_{B,B}$\label{fig:w-w-5-lowest-from-unw}]{
        \includegraphics[width=\linewidth]{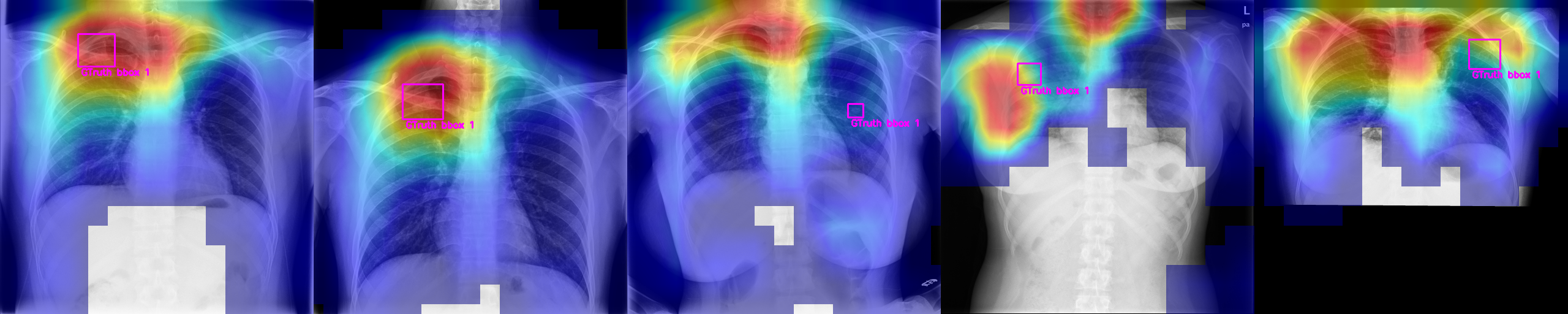}
    }
    \caption{\textsc{Saliency Maps}. In this figure, we show
    \protect\subref*{fig:unw-5-lowest-propeng} Saliency maps for the unbalanced
    model $M_U$ featuring the five cases from the TBX11K test set with the lowest
    Proportional Energy scores; \protect\subref*{fig:w-w-5-lowest-from-unw}
    respective predictions of our best balanced model $M_{B,B}$.
    Human-annotated ground-truth regions including radiological signs are
    indicated by bright magenta bounding boxes. The heatmaps (ranging from red
    to blue) indicate the contribution of different regions to the models'
    decision-making, with non-colored areas having no significant
    contribution.}
    \label{fig:prop-energy}
\end{figure*}

We have trained all five models, preserving the one reaching the lowest loss on
the validation set in the respective setups.   We then observed that all models
achieved an AUROC of 1.0 (perfect scoring) on the \emph{target} test set
(TBX11K). Subsequently, we cross-evaluated all models against the full
\emph{external} dataset (Shenzhen) to explore their generalization
capabilities.  As shown in Tab.~\ref{tab:results}, the unbalanced model $M_U$
reaches an AUROC of 0.79 when cross-evaluated, whereas the balanced model $M_B$
AUROC for a cross-dataset evaluation slightly drops to 0.73. All the other
models achieved an AUROC of at least 0.86, demonstrating higher generalization
once exposed to a larger dataset, with or without balancing.

We also evaluated each model's (human-like) interpretability using their median
Proportional Energy over all test samples with active TB of the \emph{target}
dataset (for obvious reasons there exist no bounding-box labels for the healthy
cases), listed in Tab.~\ref{tab:results}.  As can be observed, interpretability
increases as more data and balancing are introduced to the models' training
process, from below 0.1 for $M_U$ to about $0.3$ for all the saliency mapping
techniques explored when balancing is applied to both \emph{proxy} training and
\emph{target} fine-tuning. Remarkably, compared to model $M_U$, the addition of
balancing to the simple binary classifier without adding more data ($M_B$)
leads to a lower interpretability. Grad-CAM and HiResCAM perform similarly on
all models, whereas Score-CAM shows more extreme penalization on model $M_B$
with a median Proportional Energy of 0.007, and slightly better improvement on
model $M_{B,B}$ with 0.326. We note that while a perfectly aligned model would
have its median equal to 1.0, the use of bounding boxes to represent natural
radiological signs compromises this metric as such signs are rarely rectangular
in nature.

Fig.~\subref{fig:unw-5-lowest-propeng} presents five HiResCAMs visualizations with
the lowest Proportional Energy scores 
obtained by the unbalanced model $M_U$. These samples, all correctly identified
as active TB, exhibit saliency maps with notable focuses on the center of the
CXRs and outside the lung areas, \eg, in the armpit or the bottom of the image
where typically diagnostic information is absent. In
Fig.~\subref{fig:w-w-5-lowest-from-unw}, the same five samples are visualized
with the balanced model $M_{B,B}$. In the left two images, the saliency maps now
correspond to human (ground-truth) annotations, and results for the other three cases at
least exhibit a closer alignment with the patients' bodies. Hence, the
$M_{B,B}$ model not only maintains accurate classification but also
demonstrates a marked refinement in CAM localization.

\section{Conclusion}
\label{sec:conclusions}

Training deep neural networks (DNNs) in a naive manner often results in models
with undesirable biases. Contrary to previous experiences with large-scale data
\cite{zhang2024cam}, we found that training a balanced classifier on target
data decreased interpretability. However, pre-training the classifier on a
larger, related multi-objective classification task significantly mitigated
this issue and improved model transferability to an \emph{external} dataset.
Additionally, balancing classes during pre-training and fine-tuning enhanced
alignment with human expectations without compromising utility on the
\emph{target} and \emph{external} datasets.

Our study has notable findings, but also some limitations. While our models
demonstrate utility on a cross-dataset evaluation, this did not fully confirm
their generalization, suggesting the presence of residual biases. Specifically,
models pre-trained on the NIH-CXR14 dataset exhibit better generalization
capabilities. Despite observed improvements in Proportional Energy, it is
important to acknowledge that the values remain significantly lower than the
ideal score of 1. This discrepancy highlights the necessity for further
testing, particularly with more precisely annotated radiological signs, and
more diverse datasets, to better understand the limits of this metric.
Moreover, while our study demonstrates potential, it does not fully address how
these methods could be integrated into clinical settings and how they would
perform under these conditions. Finally, the human interpretability testing
methodology proposed in this study shows promise for debiasing other models
while preserving their utility.

\bibliographystyle{IEEEtran}
\bibliography{references}

\end{document}